\def\year{2021}\relax
\documentclass[letterpaper]{article} % DO NOT CHANGE THIS
\usepackage{aaai21}  % DO NOT CHANGE THIS
\usepackage{times}  % DO NOT CHANGE THIS
\usepackage{helvet}  % DO NOT CHANGE THIS
\usepackage{courier}  % DO NOT CHANGE THIS
\usepackage[hyphens]{url}  % DO NOT CHANGE THIS
\usepackage{graphicx} % DO NOT CHANGE THIS
\usepackage{natbib}  % DO NOT CHANGE THIS AND DO NOT ADD ANY OPTIONS TO IT
\usepackage{caption} % DO NOT CHANGE THIS AND DO NOT ADD ANY OPTIONS TO IT
\DeclareCaptionStyle{ruled}{labelfont=normalfont,labelsep=colon,strut=off} % DO NOT CHANGE THIS
\usepackage{algorithm}
\usepackage{algorithmic}

\usepackage{newfloat}
\usepackage{listings}
\floatstyle{ruled}
\newfloat{listing}{tb}{lst}{}
\floatname{listing}{Listing}

\usepackage{booktabs}
\usepackage{subfigure}
\usepackage{CJKutf8}
\usepackage{ragged2e,array}
\usepackage{multirow} % for borders and merged ranges
\usepackage{soul}% for underlines
\usepackage[table]{xcolor} % for cell colors
\usepackage{changepage,threeparttable} % for wide tables
\usepackage[switch]{lineno}
\usepackage{makecell}
\usepackage[title]{appendix}
\usepackage{bbding} %首先在导言区调用bbding包
\title{BiRdQA: A Bilingual Dataset for Question Answering on Tricky Riddles}
\author {
    Yunxiang Zhang,
    Xiaojun Wan \\
}
\affiliations {
    Wangxuan Institute of Computer Technology, Peking University\\
    The MOE Key Laboratory of Computational Linguistics, Peking University\\
    \texttt{\{yx.zhang,wanxiaojun\}@pku.edu.cn}
}
\usepackage{bibentry}
\begin{document}
% \linenumbers
\begin{CJK*}{UTF8}{gbsn}

\maketitle

\begin{abstract}
A riddle is a question or statement with double or veiled meanings, followed by an unexpected answer. Solving riddle is a challenging task for both machine and human, testing the capability of understanding figurative, creative natural language and reasoning with commonsense knowledge. We introduce BiRdQA, a bilingual multiple-choice question answering dataset with 6614 English riddles and 8751 Chinese riddles. For each riddle-answer pair, we provide four distractors with additional information from Wikipedia. The distractors are automatically generated at scale with minimal bias. Existing monolingual and multilingual QA models fail to perform well on our dataset, indicating that there is a long way to go before machine can beat human on solving tricky riddles. The dataset has been released to the community\footnote{\url{https://forms.gle/NvT7DfWhAPhvoFvH7}}.
\end{abstract}

\section{Introduction}\label{sec:intro}
% Motivation
In recent years, many large-scale, high-quality datasets tackle Question Answering from various angles, including extractive question \cite{rajpurkar2016squad}, multi-hop question \cite{yang2018hotpotqa}, commonsense question \cite{talmor-etal-2019-commonsenseqa} and logical question \cite{liu2020logiqa}. While the reasoning required to answer questions is going deeper and deeper, the QA community seems to neglect the importance of \textit{understanding} the question itself before \textit{reasoning} for the answer. Understanding language requires both linguistic knowledge and commonsense knowledge \cite{lobue-yates-2011-types}. We should also keep in mind that words should not always be taken at face value \cite{veale2011creative}. However, existing QA datasets are commonly sourced from Wikipedia \cite{yang2015wikiqa, rajpurkar2016squad} and news articles \cite{trischler2017newsqa}, inevitably leading to creation of superficial literal questions, which are easy to comprehend. In order to find more challenging questions with diverse linguistic styles and rich commonsense knowledge, we turn to riddles for help.
% \cite{mcgann2014riddle}.

  \begin{figure}[H]
    \includegraphics[width=\linewidth]{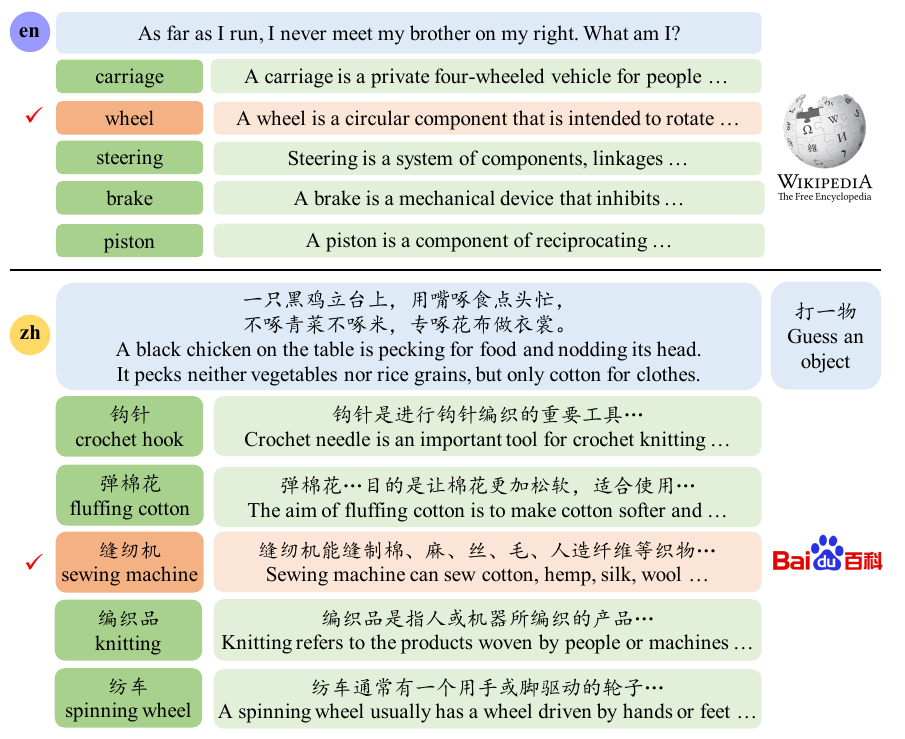}
    \caption{Examples from the English (the upper one) and Chinese (the lower one) part of BiRdQA. There are four distractors and one correct answer for a riddle and an introduction from English Wikipedia or Chinese Baidu Baike is provided for each candidate. Note that Chinese riddles have additional hints. The checkmark indicates the correct answer.}
    \label{fig:example}
  \end{figure}

A riddle is a traditional verbal expression which contains one or more descriptive elements, a pair of which may be in opposition; the referent of the elements is to be guessed \cite{robert1963definition}.  As a classical form of guessing game, riddle has been a popular part of folk literature across many countries. One of the most wide-spread riddles appears in Sophocles’s play {\it  Oedipus the King} \cite{sophocles1994oedipus}, where Oedipus is tasked with saving the city of Thebes from a Sphinx by solving her riddle, “{\it What goes on four legs in the morning, on two legs at noon, and on three legs in the evening? -- human}”. 
  While it is a matter of life and death for Oedipus to answer the riddle correctly, it is interesting to investigate the potential of AI for solving riddle, given the rise of deep learning and pretraining based methods in NLP.

Consider the English riddle shown in Figure \ref{fig:example}, ``{\it As far as I run, I never meet my brother on my right. What am I? -- wheel}''. When people initially come across the word “brother”, they naturally think of a living creature such as human or animal. But if they think deeper about the description of the brother's location (``on my right") and the  reason why it ``never meets'' its ``brother", they will realize that the riddle may be talking about an inanimate object. In fact, it is a wheel that runs in parallel with its counterpart on the opposite side of a vehicle. This typical example shows the following properties of riddle which make it challenging:

\begin{itemize}
    \item \textbf{Figurative.} A common trick of riddles is personification, which gives everyday objects human characteristics. Other types of figurative language used in riddles include metaphor, pun, and hyperbole \cite{senderovich2016review}. The usage of various figurative devices can fox existing QA models \cite{10.1145/3375547}.
    \item \textbf{Misleading.} Hard riddles are meant to play a trick on readers with an apparently irreconcilable contradiction or incongruity \cite{robert1963definition}. Aristotle commented on this characteristic of riddle in the \textit{Poetics} \cite{2013poetics} that ``the very nature indeed of a riddle is this, to describe a fact in an impossible combination of words". Commonsense reasoning and creative thinking \cite{schank1987natural} are required in order to develop a valid explanation for the seemingly impossible description of riddling object.
\end{itemize}

 We believe that riddles, as a thought provoking challenge for human, can help sharpen deeper language understanding and reasoning abilities of machines as well. On the one hand, the figurative property of riddle enables communication of far richer meanings than factual question and thus it is beneficial to introduce Figurative Language Processing (FLP) into the field of QA. On the other hand, the misleading nature of riddle puts obstacle in the way of commonsense reasoning.

Riddle is an universal art, since it exists in hundreds of different cultures \cite{taylor1977english}. Besides, many riddles and riddle objects (i.e., the object that a riddle describes) are internationally widespread \cite{aarne1918vergleichende}. Given the recent progress in multilingual QA \cite{jing2019bipar, lewis-etal-2020-mlqa}, we conduct a cross-lingual study of riddles in English and Chinese due to easy access to a large amount of available riddles online in these two primary languages. 
We present BiRdQA (\textbf{Bi}lingual \textbf{R}i\textbf{d}dle \textbf{Q}uestion \textbf{A}nswering), a collection of 6614 English riddles and 8751 Chinese riddles. BiRdQA is a multiple-choice question answering dataset, as we provide a correct answer and four distractors for every riddle. In this way, it is easy and fair to adopt accuracy as evaluation metrics to compare the performance of different QA models on our dataset. Our proposed method for automatic distractor generation allows for robust multiple-choice dataset creation at scale with minimal human effort and sufficient quality. 
To bring in more commonsense knowledge for reasoning, we also provide a brief introduction of each candidate, which is collected from Wikipedia article. Note that BiRdQA is a \textit{non-parallel} bilingual dataset, because it is extremely difficult and sometimes impossible to correctly translate riddles between different languages\footnote{For example, for the riddle ``what do you call a fruit that is never alone? -- pear", the answer makes sense because ``pear" sounds like ``peer", which means ``never alone". However in Chinese or other languages, ``pear" and ``peer" do not sound similarly.}. Figure 1 shows two selected examples of our dataset, one in English and the other in Chinese.

To assess the difficulty of BiRdQA, we conduct experiments on several baseline models. Results show a significant gap between machine and human performance, indicating that there is a lot of headroom for improvement. We also evaluate multilingual pretrained models in zero-shot cross-lingual transfer and multilingual training settings. 

The contributions of our work are summarized as follows:

\begin{itemize}
\item We construct a large challenge dataset BiRdQA in a non-trivial way. The dataset can be used for testing the question answering ability on tricky riddles. The dataset will be released and contribute to the QA and NLP community. 
\item We conduct thorough analysis on the BiRdQA dataset and benchmark the performance of pretraining based QA models on BiRdQA. Results show a significant gap between machine and human performance. 
\end{itemize}

\section{Dataset Collection}
 We cast solving riddle as a five-way multiple choice task instead of open-ended format, to allow efficient and fair evaluation of the knowledge proficiency of a model  \cite{zellers2019recognition, qiu-etal-2020-automatic}. Given a question with five candidates and their introductions, a model should select one as the correct answer. We create BiRdQA in three steps, for English and Chinese riddles independently: collecting riddles, generating distractors and collecting introductions for candidates\footnote{From now on, we use the term ``answer” for the correct solution to a riddle, ``distractors" for the four wrong choices of a question, and ``candidates" for the union of ``answer" and ``distractors" of a riddle. }. We dub English part of our dataset ``BiRdQA-en" and Chinese part ``BiRdQA-zh". We aim to maintain symmetry between the two languages with similar formats and standards of data, and comparable methods of dataset creation.

\subsection{Riddle Collection}\label{sec:riddle-collect}

We crawl 16,000+ English riddle-answer pairs from several websites\footnote{e.g., \url{www.riddlesandanswers.com}, \url{www.riddles.com}} where uniquely crafted, high-quality riddles are in abundance. We also crawl 12,000+ Chinese riddles and their answers from a single website\footnote{\url{http://www.cmiyu.com/}}. Different from English riddles, every Chinese riddle contains a hint in parenthesis (e.g., ``打一物" in Figure \ref{fig:example}) and we separate it as an additional field. Details about raw data preprocessing are described in Appendix \ref{appendix-data}.

\subsection{Distractor Generation}\label{sec:distractor-gen}
It is quite challenging to obtain high-quality distractors at scale. Asking humans to write distractors for each correct answer is expensive and time-consuming \cite{gierl2017developing, zellers2019recognition}, especially for a large-scale dataset. Moreover, it may suffer from \textit{spurious correlation}, also called \textit{annotation artifacts}: subtle patterns become prediction rules that work for the majority examples but do not hold in general \cite{gururangan2018annotation, poliak2018hypothesis, tu2020empirical}. Therefore, our motivation is that a high-quality distractor generated in an automated fashion should
\begin{itemize}
    \item be close enough to the correct answer \cite{goodrich1977distractor} to make it more plausible and adversarial, so long as it does not become another valid answer.
    \item be robust to candidate-only bias \cite{si2019does, yu2020counterfactual}, disincentivizing models from leveraging shortcuts to directly predict the correct answer without reading the question.\footnote{While it is also important for the distractors to be consistent with the semantic context of the question \cite{gao2019generating}, our proposed method is effective enough for distractor generation on this dataset, which is demonstrated empirically in  Section \ref{sec:results} later. As a result, we do not take that point into account for simplicity.} 
\end{itemize}

 Now we describe our distractor generation method in detail. Our goal is to generate four distractors for each correct answer. Two of them are designed to be similar to the correct answer in terms of the cosine distance of word embedding\footnote{If the correct answer of a riddle does not have pretrained embedding, we simply remove this riddle.}, which we refer to as \textit{global distractors}. The other two distractors are drawn from the correct answers of other questions \cite{zellers2019recognition}, so as to overcome the issue of candidate-only bias. We use \textit{local distractors} to refer to these two distractors. For example, for the two riddles in Figure \ref{fig:example}, ``carriage", ``brake", ``弹棉花" (fluffing cotton), and ``纺车" (spinning wheel) are local distractors, while the rest are global ones.

\paragraph{Global Distractor} For each English riddle, we use the correct answer as the query and retrieve the top-2 most similar words or phrases from the pretrained Sense2Vec \cite{Trask2015sense2vecA, pan-etal-2021-zero}. For each Chinese riddle, we use the word embedding provided by \citet{song-etal-2018-directional}. 

\paragraph{Local Distractor} For each language, we build an answer pool with all distinct correct answers of riddles. For each riddle, we obtain the top-2 most similar correct answers from that answer pool. The similarity is also measured by the cosine distance of word embedding and the choices of pretrained embedding for English and Chinese are the same as \textit{global distractors}.

\paragraph{Invalid Distractor Replacement} To avoid the case that the generated distractor becomes another correct answer or repeats with other distractors, we define rules to ensure that the distractor has less lexical overlap with the correct answer \cite{pan-etal-2021-zero} and other distractors of the same riddle. Specifically, two candidates are considered as lexically overlapped if they share at least a same word in English (e.g., ``horse" and ``wild horse") or a same character in Chinese (e.g., ``蘑菇" and ``平菇"). We iteratively replace the problematic distractor with the next best (i.e., most similar to correct answer) generated one until it is valid. We generate local distractors before global ones because the larger search space of global distractors allows more alternatives for replacement. We provide a detailed quality analysis of the generated distractors in Section \ref{sec:analysis}.

\subsection{Candidate Introduction Collection}

Wikipedia has been widely used as an important knowledge source for Question Answering \cite{yang2015wikiqa, chen-etal-2017-reading, liu-etal-2020-rikinet} and Commonsense Reasoning \cite{storks2019commonsense, lv2020graph}. We provide a brief Wikipedia-based introduction of each candidate to help models solve riddles with commonsense knowledge. For Chinese riddles we use Baidu Baike\footnote{\url{https://baike.baidu.com/}}, the Chinese equivalent to Wikipedia, because it has richer Chinese entries\footnote{We use the word ``Wikipedia" to refer to both English Wikipedia and Baidu Baike (Chinese online encyclopedia) hereafter.}. We match each candidate to its corresponding Wikipedia entry and use the introductory part of the article (i.e., the content before the first section) as its introduction. We define that a candidate is matched to a Wikipedia entry if its name is same as the title of the Wikipedia entry or can be redirected to that entry (e.g., ``stone" $\rightarrow$ ``Rock (geology)"\footnote{\url{https://en.wikipedia.org/wiki/Rock_(geology)}}). To resolve disambiguation issues (e.g., ``nail" may refer to ``Nail (anatomy)"\footnote{\url{https://en.wikipedia.org/wiki/Nail_(anatomy)}}, ``Nail (beak)"\footnote{\url{https://en.wikipedia.org/wiki/Nail_(beak)}}, etc.), we select the entry 
whose introduction has the highest similarity with the question. We calculate the similarity with Sentence-BERT \cite{reimers-2019-sentence-bert}. Specially, if the correct answer of a riddle does not have Wikipedia entry, we simply remove this riddle. If a generated distractor does not have linked Wikipedia entry, we iteratively replace it with the next best (i.e., most similar to correct answer) generated one until it has one. In this way, we ensure that all candidates in our dataset are accompanied by introductions. We evaluate the quality of collected Wikipedia introductions in Section \ref{sec:analysis}. 

Overall, we collect 6614 English and 8751 Chinese riddles with their answers, distractors and introductions. We observe that there are multiple riddles describing the same object in different ways and thus sharing identical answer (Section \ref{sec:analysis}). If they appeared in both training and development/test set, the machine might memorize the correct answer rather than reason it out \cite{zellers2019recognition, talmor-etal-2019-commonsenseqa}. Therefore, when splitting the dataset into training/development/test set, we ensure that each of the three sets has disjoint correct answers.

\section{Dataset Analysis}
\label{sec:analysis}

\begin{table}
  \centering
\begin{tabular}{l|cc}
\toprule
Measurement & BiRdQA-en & BiRdQA-zh \\
\midrule
\# Training examples & 4093  & 5943 \\
\# Validation examples & 1061  & 1042 \\
\# Test examples & 1460  & 1766 \\
\# Total examples & 6614  & 8751 \\
\midrule
\# Avg question tokens & 29.74 & 8.23 \\
\# Distinct question tokens & 15043 & 18298 \\
\# Distinct hints & N/A   & 690 \\
\# Distinct answers & 3057  & 4750 \\
\# Distinct candidate tokens & 6593  & 14944 \\
\# Avg introduction tokens & 231.64 & 90.93 \\
\bottomrule
\end{tabular}%
\caption{Statistics of BiRdQA.}\label{tab:stat}%
\end{table}%

Table \ref{tab:stat} describes the key statistics of BiRdQA. Tokens in Chinese are referred to as words segmented by jieba\footnote{\url{https://github.com/fxsjy/jieba}} toolkit. Generally speaking, English riddles and candidate introductions are much longer than Chinese ones, while Chinese riddles have more diverse tokens in questions and candidates. The hints in Chinese riddles do not vary a lot, because they only tell the category of the answer.

\begin{table}
  \centering
\begin{tabular}{lcc|lcc}
\toprule
\multicolumn{1}{l}{Answer} & \# en & \# zh   & \multicolumn{1}{l}{Answer} & \# en & \# zh \\
\midrule
\multicolumn{6}{c}{BiRdQA-en} \\
\midrule
\multicolumn{1}{l}{water} & 45 &  0 & \multicolumn{1}{l}{moon} & 30 & 0 \\
\multicolumn{1}{l}{clock} & 43 & 3 & \multicolumn{1}{l}{sun} & 29 & 0 \\
\multicolumn{1}{l}{shadow} & 38  & 0 & \multicolumn{1}{l}{heart} & 26 & 0 \\
\multicolumn{1}{l}{egg} & 33 &  0  & \multicolumn{1}{l}{tree} & 26 & 0 \\
\multicolumn{1}{l}{fire} & 30 & 0  & \multicolumn{1}{l}{candle} & 25 & 10 \\
\midrule
\multicolumn{6}{c}{BiRdQA-zh} \\
\midrule
\makecell[l]{玉米 \\ corn} & 9 & 36    & \makecell[l]{公鸡  \\ rooster} & 8 & 25 \\
\makecell[l]{蜻蜓 \\ dragonfly} & 0 & 30    & \makecell[l]{蚊子 \\ mosquito} & 6 & 24 \\
\makecell[l]{老鼠 \\ mouse} & 9 & 27    & \makecell[l]{棉花 \\ cotton} & 2 & 22 \\
\makecell[l]{青蛙 \\ frog} & 9 & 27    & \makecell[l]{狗 \\ dog} & 11 & 21 \\
\makecell[l]{燕子 \\ swallow} & 1 & 25    & \makecell[l]{猫 \\ cat} & 15 & 20 \\
\bottomrule
\end{tabular}%

      \caption{Frequencies of 10 most common answers and their counterparts in English and Chinese riddles of BiRdQA.}
  \label{tab:counter}%
\end{table}%

\paragraph{Quality of Distractors}
Although we have reduced lexical overlap between the correct answer and distractors, we cannot completely avoid the problem that the automatically generated distractor becomes another correct answer for the riddle coincidentally (i.e., multiple candidates are equally valid). 
For example, in a Chinese riddle, the distractor ``润色先生" (inkstone) is just another name for the answer ``砚" (inkstone). Similarly, in an English riddle, the distractor ``acne scar" has the same meaning as the  answer ``zit". However, we observe that such coincidences due to synonym rarely happen: we randomly sample 100 English and 100 Chinese riddles from the development set, among which we only observe 2 such cases for each group respectively. Therefore, we leave them as  noise in the dataset. This shows that eliminating lexical overlap between generated distractors and the correct answer is an effective strategy to ensure distractor quality.

\paragraph{Quality of Introductions}
We manually check whether the Wikipedia introduction of a correct answer actually matches the answer itself\footnote{We do not check the matching quality for the distractor introductions because the ambiguous Wikipedia entries for a distractor are still wrong for the question.} by inspecting the 100 English and 100 Chinese random samples from the development set. We only observe 3 and 6 such cases, respectively, mainly due to the lack of corresponding Wikipedia entries. As Wikipedia introduction only serves as an auxiliary source of information in our dataset, we consider this quality as acceptable.

\paragraph{Distribution of Answers}

We notice that there are multiple riddles describing the same object from different perspective within a language or even across languages and thus sharing the same answer \cite{taylor1977english}, though they are not duplicate riddles (we have already removed duplicate riddles in Appendix \ref{appendix-data}). For example, for the riddles ``\textit{I'm often running yet I have no legs.  You need me but I don't need you. What am I?}" and ``\textit{When it rains, I never get any wetter. What am I?}", both of them have ``water" as their answers. Table \ref{tab:counter} shows the frequencies of the most common answers in BiRdQA. We observe that common answers in English have higher frequencies than Chinese, and English riddles have less distinct answers (Table \ref{tab:stat}). This indicates that Chinese answers are more diverse than English. It is also interesting that common answers for Chinese riddles are mostly living things such as animals, while non-living things are more popular in English riddles. The overlap of the riddle objects between English and Chinese makes it easier for models to learn riddles in one language and transfer the knowledge to another language, which is further described in the cross-lingual setting in Section \ref{sec:methods}.

\begin{table}
  \centering
    \begin{tabular}{l|cc}
    \toprule
    Figure of Speech & \# en  & \# zh  \\
    \midrule
    Hyperbole & 3     & 3 \\
    Metaphor & 6     & 12 \\
    Personification & 18    & 11 \\
    Pun   & 11    & 16 \\
    \midrule
    Used  & 31    & 44 \\
    \bottomrule
    \end{tabular}%
    \caption{Figures of Speech and their frequencies in 100 English and 100 Chinese sampled riddles.}\label{tab:figurative}%
\end{table}%
\paragraph{Figures of Speech in Riddles}

An important property of the riddle is the usage of figures of speech (Section \ref{sec:intro}). We randomly sample 100 English and 100 Chinese riddles from the development set, and manually identify the existence and types of figures of speech. As each example can be annotated with multiple figures of speech, the total frequency in Table \ref{tab:figurative} does not sum to ``Used", which means that a riddle uses at least one type of figure of speech. Table \ref{tab:figurative} shows that a significant proportion of riddles in English and Chinese leverage figurative devices to confuse readers about their true meanings. 

\section{Methods}\label{sec:methods}
We conduct experiments with multiple pretraining models on BiRdQA under monolingual, cross-lingual and multilingual settings \cite{jing2019bipar}. 
\begin{itemize}
    \item \textbf{Monolingual.} We use data in the same language for training and evaluating models (i.e., en $\rightarrow$ en, zh $\rightarrow$ zh).
    \item \textbf{Cross-lingual.} 
    We test performance in zero-shot cross-lingual transfer learning, where a multilingual pretrained model is fine-tuned on one source language and evaluated on a different target language (i.e., en $\rightarrow$ zh, zh $\rightarrow$ en).
    \item \textbf{Multilingual.} We directly mix training instances of the two languages into a single training set and build a single QA model to handle bilingual riddles in BiRdQA (i.e., en+zh $\rightarrow$ en, en+zh $\rightarrow$ zh). 
\end{itemize}
We also conduct additional experiments to  investigate the influence of adding introductions (``w/ introduction") or hints (``w/ hint"), removing questions from inputs (``w/o question"), and transfer learning with CommonsenseQA \cite{talmor-etal-2019-commonsenseqa} (``Train + CQA", ``Train = CQA") on one or multiple settings, which are detailed in Section \ref{sec:results} later\footnote{To clarify, we only use introductions and hints in additional experiments.}.

\subsection{Baseline Models}
Below we describe the specific English models and Chinese models for monolingual setting, as well as the multilingual models for cross-lingual and multilingual settings.
\paragraph{English Models}
  We test the performance of BERT \cite{devlin2019bert} and its variants, RoBERTa \cite{liu2019roberta} and ALBERT \cite{lan2019albert}.  Following the standard multiple-choice QA setup for pretrained language models \cite{devlin2019bert}, we treat the question as \textit{A} and each candidate as \textit{B}, before further linearizing them into \textit{[CLS] A [SEP] B [SEP]} for encoding. The \textit{[CLS]} token embedding is used to produce a score for each candidate, which is then passed through a softmax layer for final prediction. We also experiment with UnifiedQA \cite{khashabi2020unifiedqa}, the state-of-the-art QA model for many QA benchamarks. We feed the question and all five candidates together to UnifiedQA so that it can choose among the candidates instead of judging them independently like BERT.

\paragraph{Chinese Models}
We experiment with popular Chinese pretrained models, including Chinese BERT \cite{devlin2019bert}, BERT-wwm, RoBERTa-wwm \cite{cui2019pre} and ERNIE \cite{zhang-etal-2019-ernie}. BERT-wwm and RoBERTa-wwm are models pretrained with Whole Word Masking (WWM). ERNIE incorporates knowledge graphs (KGs) to enhance language representation with external knowledge. We adopt the fine-tuning procedure similar to English BERT.

\paragraph{Multilingual Models}
Multilingual pretrained language models are able to perform cross-lingual zero-shot learning on multilingual NLU tasks such as XNLI \cite{conneau-etal-2018-xnli} and MLQA \cite{lewis-etal-2020-mlqa}. We test the performance of multilingual BERT (mBERT) \cite{devlin2019bert} and XLM-R \cite{conneau2019unsupervised} on BiRdQA. We adopt the fine-tuning procedure similar to English BERT.

\paragraph{Input Format}
For additional experiments regarding the influence of adding introduction/hint, the \textit{B} in \textit{[CLS] A [SEP] B [SEP]} for BERT-style input is instead the concatenation of the candidate and its introduction/hint, while the \textit{A} is still the question.

\subsection{Evaluation Metric}
We evaluate the model performance with two metrics, accuracy and mean reciprocal rank (MRR), which show how well the model makes inference and ranks the candidates, respectively. The model ranks the candidates by their softmax prediction probabilities, except for UnifiedQA. Because UnifiedQA is an end-to-end generative model, we cannot apply MRR metric to it.

\subsection{Experimental Setup}\label{sec:exp-setup}
We use Huggingface\footnote{\url{https://huggingface.co/models}} implementations for all the baseline models. Due to limitation of computational resource, we restrict the input length to 256 tokens for all models except 150 for UnifiedQA.  All hyper-parameters are decided by the model performance on the development set. For cross-lingual experiment, model selection is constrained to be strictly zero-shot, using only source language development data to pick hyperparameters  \cite{lewis-etal-2020-mlqa}. 
To keep symmetric input formats of English and Chinese riddles,
we do not use the hints in Chinese riddles because they do not exist in English riddles. We further investigate the effect of adding hints with additional experiments.

\subsection{Human Performance Evaluation}
We evaluate human performance on BiRdQA under monolingual setting. We employ three bilingual post-graduate students to independently answer 100 Chinese riddles and 100 English riddles randomly sampled from the test sets. They are instructed not to directly search for the answers of riddles online. We then calculate the average accuracy of the three human workers as the final human performance on our dataset.

\section{Results}\label{sec:results}

Tables \ref{tab:en-result} and \ref{tab:zh-result} show monolingual experiment results on BiRdQA-en and BiRdQA-zh, respectively. Table \ref{tab:en-zh-result} describes cross-lingual and multilingual results. We refer our readers to Appendix \ref{appendix:exp-result} for more comprehensive results. The human accuracy is 81.33\% for English and 87.67\% for Chinese\footnote{Because human performances are evaluated on a subset of 100 random samples from the test set, they cannot be strictly compared with the model performances evaluated on the whole test set.}, which indicates that it is not difficult for human testees to distinguish the correct answer from other distractors, though human may struggle to answer the riddles ``from scratch" (i.e., no candidates are provided). In contrast, all of the evaluated QA models perform much worse than human, indicating that theses methods are relatively weak in solving riddle and that there is ample room for improvement in future.

\begin{table}
  \centering
    \begin{tabular}{l|cc}
    \toprule
    \multirow{2}[4]{*}{Model} & \multicolumn{2}{c}{BiRdQA-en} \\
\cmidrule{2-3}          & Dev   & Test \\
    \midrule
    Random Guess & 20.00 / 0.4567 & 20.00 / 0.4567 \\
    \midrule
    BERT-Base & 48.82 / 0.6834 & 41.92 / 0.6404 \\
    \quad w/ introduction & 51.46 / 0.6949 &	44.38 / 0.6609\\
    \quad w/o question & 29.59 / 0.5519 &	21.44 / 0.4835 \\
    \quad Train + CQA & 46.56 / 0.6771 &	44.86 / 0.6599\\
    \quad Train = CQA & 42.41 / 0.6463 &	38.29 / 0.6164 \\\cmidrule{1-3}
    BERT-Large & 46.65 / 0.6735 & 44.25 / 0.6519 \\
    RoBERTa-Large & 47.31 / 0.6749 & 45.21 / 0.6653 \\
    ALBERT-XXL & 63.52 / 0.7856 & 58.70 / 0.7590 \\
    \quad  Train + CQA & 67.11 / 0.8045 &	\textbf{64.79 / 0.7978} \\
    UnifiedQA (T5-Large) & \textbf{67.20}  & 62.60  \\
    \midrule
    Human & -     & 81.33  \\
    \bottomrule
    \end{tabular}%
    \caption{Results (accuracy/MRR) of English models on the development and the test data of BiRdQA-en. CQA means CommonsenseQA \cite{talmor-etal-2019-commonsenseqa}. Human performance is tested on a subset of 100 random samples.}\label{tab:en-result}%
\end{table}%

\paragraph{English vs. Chinese}
In monolingual setting, the best performance on BiRdQA-en is 62.60\% by UnifiedQA (without additional data), higher than that on BiRdQA-zh, which is 59.29\% by ERNIE. However, English results are almost worse than the corresponding Chinese results for the same type of model (e.g., BERT-Base vs. BERT-base-chinese). We also observe similar trends in the multilingual setting. However, this does not necessarily mean that English riddles are generally harder than Chinese riddles, since there are less training samples for English riddles and also English answers are less diverse than Chinese ones (Section \ref{sec:analysis}). In another word, models tend to learn less knowledge in English. 
In future, we plan to increase diversity of English riddle and make the dataset more balanced between the two languages. We provide an error analysis in Appendix \ref{appendix:error}.

\paragraph{Monolingual vs. Multilingual} 
In monolingual training setting, monolingual model (BERT) can outperform its multilingual counterpart (mBERT), which is called \textit{the curse of multilinguality} \cite{conneau2019unsupervised}. We also observe that the multilingual training (e.g., en+zh $\rightarrow$ en) substantially improves the performance comparing with the monolingual training (e.g., en $\rightarrow$ en), especially for English riddles. It shows that multilingual models are able to transfer knowledge from source language to improve performance on different target language on our dataset.

\paragraph{Comparing Different QA Models} 
In monolingual English setting (Table \ref{tab:en-result}), UnifiedQA performs the best (without additional data) since it can see all the candidates in the input sequence at the same time rather than consider them independently like BERT. Another possible reason is that UnifiedQA has been pretrained on other QA dataset, making it possible to transfer external knowledge and skills to BiRdQA. In monolingual Chinese setting (Table \ref{tab:zh-result}), ERNIE achieves the best accuracy surprisingly, given that it has less parameters than RoBERTa-wwm-ext-large. The difference of the domains of pretraining text may account for this phenomenon. ERNIE is trained on larger data other than Wikipedia, including text from online forum, which will be more useful on casual text like riddles \cite{cui2019pre}. We note that higher accuracy does not necessarily means higher MRR, especially when accuracies of two models are very close to each other. For example, in Table \ref{tab:zh-result}, ERNIE beats RoBERTa on accuracy but is left behind on MRR, which shows that RoBERTa is better at ranking candidates.
We also observe that larger models consistently perform better than smaller models. In cross-lingual setting, mBERT is the best one for zero-shot transfer, which is different from the case on MLQA \cite{lewis-etal-2020-mlqa}. While in multilingual setting, XLM-R-Large outperforms other models on both English and Chinese test sets.

\begin{table}
  \centering
\begin{tabular}{l|cc}
\toprule
\multirow{2}[4]{*}{Model} & \multicolumn{2}{c}{BiRdQA-zh} \\
\cmidrule{2-3}      & Dev   & Test \\
\midrule
Random Guess & 20.00 / 0.4567 & 20.00 / 0.4567 \\
\midrule
BERT-base-chinese  &  53.45 / 0.7099 & 55.10 / 0.7210 \\
\quad w/ introduction & 55.95 / 0.7244 &      54.08 / 0.7102 \\
\quad w/o question & 28.79 / 0.5413 &	22.48 / 0.4954 \\
\quad w/ hint & 54.32 / 0.7162 &	54.08 / 0.7139\\\cmidrule{1-3}
BERT-wwm-ext & 58.25 / 0.7421     &      57.02 / 0.7353 \\
RoBERTa-wwm-ext-large &  60.17 / \textbf{0.7563}      &     58.04 / 0.7415 \\
ERNIE & \textbf{60.65} / 0.7550      &     \textbf{59.29} / \textbf{0.7479} \\
\midrule
Human & - &  87.67 \\
\bottomrule
\end{tabular}%
    \caption{Results (accuracy/MRR) of Chinese models on the development and the test data of BiRdQA-zh. Human performance is tested on a subset of 100 random samples. }\label{tab:zh-result}%
\end{table}%

\begin{table*}
\centering
\begin{tabular}{l|cc|cc|ccc}\toprule
Setting &\multicolumn{4}{c}{Cross-lingual} &\multicolumn{2}{|c}{Multilingual} \\\cmidrule{1-7}
Train &\multicolumn{2}{c|}{en} &\multicolumn{2}{c|}{zh} &\multicolumn{2}{c}{en+zh} \\\cmidrule{1-7}
Test &en &zh &zh &en &en &zh \\\cmidrule{1-7}
mBERT &36.30 / 0.6003 &\textbf{40.60 / 0.6188} &50.17 / 0.6845 & 38.97 / 0.6144 &40.14 / 0.6246 &50.51 / 0.6906 \\
\quad w/ introduction & 42.40 / 0.6467 & 30.12 / 0.5521 & 47.85 / 0.6677 & \textbf{41.23 / 0.6359} & 43.08 / 0.6505  & 48.75 / 0.6771 \\\cmidrule{1-7}
XLM-R-Base &28.77 / 0.5408 &31.20 / 0.5552 &46.49 / 0.6646 &25.21 / 0.5181 &33.77 / 0.5783 &48.41 / 0.6769 \\
XLM-R-Large &39.73 / 0.6249 &33.01 / 0.5651 &57.25 / 0.7319 &38.15 / 0.6099 &\textbf{43.97 / 0.6591} &\textbf{57.13 / 0.7335} \\
\bottomrule
\end{tabular}
\caption{Results (accuracy/MRR) of multilingual models under cross-lingual and multilingual settings of BiRdQA. We only mark the best performances in cross-lingual (en $\rightarrow$ zh, zh $\rightarrow$ en) and multilingual (en+zh $\rightarrow$ en/zh) experiments as bold.}\label{tab:en-zh-result}
\end{table*}

\paragraph{Influence of Additional Information} Wikipedia introductions and hints (only for Chinese riddles) are auxiliary information in our dataset and they may be useful to models. We evaluate the impact of appending Wikipedia introduction to each candidate for input.  We find that adding Wikipedia introduction (``w/ introduction" in Tables \ref{tab:en-result} and \ref{tab:zh-result} and \ref{tab:en-zh-result}) can boost performance on English riddles across the three experimental settings, but cannot improve performance on Chinese ones, unexpectedly. We conjecture that this is because solving Chinese riddles requires deeper information of candidates, such as the meaning of each character\footnote{Unlike English, each Chinese character has its own meaning and the combination of character meanings does not necessarily equal to the word meaning.}, rather than factual/literal descriptions from Wikipedia.
We also observe that using hints in Chinese riddles (``w/ hint" in Table \ref{tab:zh-result}) does not
lead to higher performance, indicating that they do not carry a lot of useful information. For example, if all five candidates are animals, the hint ``打一动物" (guess an animal) will be useless. This is also correlated to the fact that the hint is a less diverse feature in our dataset -- Table \ref{tab:stat} shows that there are only 690 distinct hints across all Chinese riddles.

\paragraph{Investigating Candidate-only Bias}
We conduct ablation experiments to measure candidate-only bias by removing question from input (``w/o question" in Table \ref{tab:en-result} and \ref{tab:zh-result}). 
We observe that models can only make nearly random guess (20\%) without the question, demonstrating the effectiveness of local distractors (Section \ref{sec:distractor-gen}) to prevent machine from learning spurious correlations between distractors and correct answers. 

\paragraph{Transfer Learning with CommonsenseQA}
BiRdQA-en is similar to CommonsenseQA \cite{talmor-etal-2019-commonsenseqa} in terms of commonsense reasoning. We conduct transfer learning experiments to gauge the degree of overlap of necessary knowledge between these two datasets. We append the training set of CommonsenseQA to BiRdQA (``Train + CQA" in Table \ref{tab:en-result}), and observe an improvement of performance from 41.92\% to 44.86\% on BERT-Base model and achieve the highest performance so far (64.79\%) on ALBERT-XXL. Besides, we fine-tune a model on the data of CommonsenseQA only (``Train = CQA" in Table \ref{tab:en-result}), and achieve a zero-shot transfer performance of 38.29\% with BERT-Base on BiRdQA-en, which indicates that there exists an overlap in terms of commonsense knowledge for solving problems in both BiRdQA and CommonsenseQA.

\section{Related Work}

To the best of our knowledge, RiddleSense \cite{lin-etal-2021-riddlesense} is the most similar dataset to ours. It contains 5.7k English riddles and also comes in a multiple-choice format. However, we argue that our work is different in that 
1) BiRdQA contains bilingual riddles and thus can facilitate a cross-lingual study of riddles.
2) BiRdQA provides Wikipedia introduction for each candidate to promote future research on commonsense reasoning for riddles.
3) We generate distractors in an automated fashion with quality guarantee and less human efforts. It can work for English, Chinese and other languages as long as pretrained word embeddings are available.
4) The bias in BiRdQA-en is much weaker than RiddleSense. A BERT-Base model simply achieves an accuracy of 42.21\% even without question input on RiddleSense, compared with only 21.44\% on BiRdQA-en.
5) RiddleSense makes use of ConceptNet for distractor generation, which is potentially unfair for non-ConceptNet-based models and thus restricts future research directions.
% \footnote{The leaderboard of CommonsenseQA, which also generate distractors with ConceptNet, does not accept submissions of models that make use of ConceptNet.}
We further compare BiRdQA-en with RiddleSense in Appendix \ref{appendix:comp}.

 \citet{tan2016solving} has studied on the solution and generation of Chinese \textit{character} riddle, 
 which present challenge regarding the structure of Chinese character rather than the meaning of the answer. 
But BiRdQA-zh explicitly excludes this type of Chinese riddle (Appendix~\ref{appendix-data}).

There are also commonsense-based QA datasets such as COSMOS QA \cite{huang2019cosmos} and CommonsenseQA \cite{talmor-etal-2019-commonsenseqa}. However, solving riddle requires more than just commonsense knowledge. While normal commonsense questions navigate the readers to the right direction for the answer, riddles tend to mislead them on purpose. Besides, the machine needs to understand the figurative and ambiguous effect of the riddle language. Therefor, riddle is a great benchmark for higher-order understanding of natural language and broader intelligence. 

There is a trend for multilingual QA in recent years, with the release of BiPaR \cite{jing2019bipar}, XQuAD \cite{Artetxe:etal:2019}, MLQA \cite{lewis-etal-2020-mlqa} and other cross-lingual QA datasets. While translation from English dataset is a common approach to the creation of multilingual dataset, we build BiRdQA independently yet symmetrically in English and Chinese, due to the difficulty of riddle translation.

\section{Conclusion}
In this paper, we introduce BiRdQA, a large-scale,  bilingual multiple-choice question answering dataset to facilitate the development of QA systems capable of solving tricky riddles. The huge gap between the human and machine leaves much room for improvement.
% We would like to point out some directions for future research on BiRdQA. 
% It is an open-end question for how to collect high-quality distractors, especially incorporating information of questions, to challenge the model. 
In future work, we plan to extend BiRdQA with riddles in other languages and incorporate figurative language understanding into riddle solving.
% In order not to be fooled by figurative effect of riddles, incorporating figurative language understanding into riddle solving are promising for improvements on BiRdQA. 
We hope that BiRdQA will stir more research for question answering on riddles.

\section*{Acknowledgments}
This work was supported by Hi-Tech R\&D Program of China (No. 2018YFB1005100), Beijing
Academy of Artificial Intelligence (BAAI) and State Key Laboratory of Media Convergence Production Technology and Systems. Xiaojun Wan is the corresponding author.
% We appreciate the anonymous reviewers for their helpful comments. 

\bibliography{aaai21.bib}

\begin{appendices}
\section{Data Preprocessing}\label{appendix-data}
\paragraph{English Riddle}
We take the following steps to clean the raw data of English riddles.

\begin{enumerate}
    \item We remove prefixes like ``I am ...", ``A ...", ``The ..." in the answer to keep only the riddle object. For example, the answer “{\it I am money}” turns into “{\it money}”.
    \item We remove riddles with more than three words in their answers, since longer answer usually presents too complicated scenario for machines. For instance, we remove the riddle: “{\it A cowboy rides into town on Friday, stays for three days, then leaves on Friday. How did he do it? — His horse’s name was Friday.}” In another word, we focus on riddles asking “what” questions instead of “how” or “why” questions.
    \item We remove alphabet riddles whose answer is a single English letter.
    \item We manually remove math riddles since the ability of mathematical reasoning \cite{ling2017program,pikekos2021measuring} is beyond the scope of this study. An example of math riddle is “{\it Eggs are \$0.12 a dozen. How many eggs can you get for a dollar? — 100 eggs.}” 
    \item We remove duplicate riddles collected from different sources. Two riddles are considered as the same if they have the same answer and the cosine similarity between their questions exceeds certain threshold\footnote{We set the threshold as 0.8 based on manual inspection of the data.}. We calculate the sentence similarity with Sentence-BERT \cite{reimers2019sentence}.
\end{enumerate}

\paragraph{Chinese Riddle}

We exclude Chinese \textit{character} riddles\footnote{They are manually separated  from other riddles on the source website.}, which have single Chinese characters as solutions and describe the structures of characters \cite{tan2016solving}. Instead we focus on common Chinese riddles which imply the \textit{meaning} of the answer. 
We also observe that the answers of Chinese riddles are neat and concise enough, requiring no extra process of data cleaning. Since the Chinese riddles are collected from a single source, we do not need to remove duplicate riddles either. 

\section{More Experimental Results}\label{appendix:exp-result}

We show more results on BiRdQA-en in Table \ref{tab:more-en-result} and BiRdQA-zh in Table \ref{tab:more-zh-result}. Due to limitation of computational resources, we cannot run experiment on ALBERT-XXL if the input is longer with introduction. For UnifiedQA, the concatenation of five candidates and their introductions is
often too long to be fed into the model. We observe trends that adding introduction is beneficial for English riddle solving but the improvements on Chinese riddles are negative. Fine-tuning on CommonsenseQA \cite{talmor-etal-2019-commonsenseqa} can greatly improve performance on BiRdQA-en. Adding hints does not help with prediction on BiRdQA-zh.

\begin{table*}
\centering

\scriptsize
\begin{tabular}{l|cc|cc|cc|cc}\toprule
\multirow{2}[4]{*}{Model} &\multicolumn{2}{c|}{BiRdQA-en} &\multicolumn{2}{c|}{w/ introduction} &\multicolumn{2}{c|}{Train = CQA} &\multicolumn{2}{c}{Train + CQA} \\\cmidrule{2-9}
&Dev &Test &Dev &Test &Dev &Test &Dev &Test \\\cmidrule{1-9}
BERT-Base &48.82 / 0.6834 &41.92 / 0.6404 &\textbf{51.46 / 0.6949} &44.38 / 0.6609 &42.41 / 0.6463 &38.29 / 0.6164 &46.56 / 0.6771 &44.86 / 0.6599 \\
BERT-Large &46.65 / 0.6735 &44.25 / 0.6519 &49.58 / 0.6868 &\textbf{46.30 / 0.6728} &41.28 / 0.6370 &39.04 / 0.6260 &49.76 / 0.6987 &50.41 / 0.6993 \\
RoBERTa-Large &47.31 / 0.6749 &45.21 / 0.6653 &45.90 / 0.6623 & 42.19 / 0.6480 &47.03 / 0.6715 &47.26 / 0.6751 &55.98 / 0.7384 &54.79 / 0.7302 \\
ALBERT-XXL &63.52 / 0.7856 &58.70 / 0.7590 &- &- &\textbf{55.42 / 0.7305} &\textbf{54.66 / 0.7277} &\textbf{67.11 / 0.8045} &\textbf{64.79 / 0.7978} \\
UnifiedQA (T5-Large) &\textbf{67.20} &\textbf{62.60} &- &- &48.92 &48.42 &65.22 &63.22 \\
\bottomrule
\end{tabular}
\caption{Results (accuracy/MRR) of English models on the development and the test data of BiRdQA-en.}\label{tab:more-en-result}
\end{table*}

\begin{table*}
\centering
\scriptsize
\begin{tabular}{l|cc|cc|cc}\toprule
\multirow{2}[4]{*}{Model} &\multicolumn{2}{c|}{BiRdQA-zh} &\multicolumn{2}{c|}{w/ introduction} &\multicolumn{2}{c}{w/ hint} \\\cmidrule{2-7}
&Dev &Test &Dev &Test &Dev &Test \\\cmidrule{1-7}
BERT &53.45 / 0.7099 &55.10 / 0.7210 &55.95 / 0.7244 &54.08 / 0.7102 &53.26 / 0.7114 &50.62 / 0.6954 \\
BERT-wwm-ext &58.25 / 0.7421 &57.02 / 0.7353 &57.77 / 0.7380 &55.78 / 0.7213 &56.33 / 0.7328 &56.34 / 0.7276 \\
RoBERTa-wwm-ext-large &60.17 / \textbf{0.7563} &58.04 / 0.7415 &53.26 / 0.7073 &51.19 / 0.6983 &\textbf{58.16 / 0.7456} &\textbf{58.15 / 0.7406} \\
ERNIE &\textbf{60.65} / 0.7550 &\textbf{59.29} / \textbf{0.7479} &\textbf{62.48 / 0.7674} &\textbf{56.23 / 0.7285} &57.01 / 0.7362 &57.53 / 0.7418 \\
\bottomrule
\end{tabular}
\caption{Results (accuracy/MRR) of Chinese models on the development and the test data of BiRdQA-zh.}\label{tab:more-zh-result}
\end{table*}

\section{Error Analysis}\label{appendix:error}
Table \ref{tab:error} shows two mistaken examples from English and Chinese part of our dataset, respectively. The lower the prediction probability is, the less confident the model is for its prediction. The first example of English riddle uses personification to present a seemingly impossible scenario (``doesn't sleep", ``never eats") regarding the river bed and river mouth. The distractor ``ocean" is invalid because ocean does not have ``mouth", though it does have its ``bed", or the so-called seafloor. As for the second English riddle, it requires \textit{creative thinking} \cite{schank1987natural} to come up with the rationale that when a person says the word ``silence", the silence is broken. The third example is about a Chinese riddle which uses several metaphors (e.g., ``吃肉", ``eat meats") to describe the kitchen knife. The ``but never" clue also marks the misleading property of riddle. The fourth example is very difficult for machine because it describes the meaning of each \textit{character} in the answer word rather than the meaning of the whole word (``云长" $\rightarrow$ ``关"\footnote{``云长" (Yunchang) is the courtesy name for Guan Yu (``关羽").}, ``想起" $\rightarrow$ ``怀"\footnote{Both of them mean ``recall".}, ``玄德" $\rightarrow$ ``备"\footnote{``玄德" (Xuande) is the courtesy name for Liu Bei (``刘备").}, ``来" $\rightarrow$ ``至"\footnote{``来" serves as a complement in this riddle but also has the same meaning of ``come" with ``至".}), which is an uncommon technique in English riddles. Note that this is still not a Chinese \textit{character} riddle because it is not about the structure of the single Chinese character of the answer. Overall, the figurative and misleading properties of riddles present a novel challenge for existing  QA models.

\begin{table*}
\centering
\scriptsize
\begin{tabular}{l|>{\RaggedRight}p{1.5cm}|>{\RaggedRight}p{1.5cm}|>{\RaggedRight}p{1.5cm}|>{\RaggedRight}p{1.5cm}|>{\RaggedRight}p{1.5cm}}\toprule
Riddle &\multicolumn{5}{>{\RaggedRight}p{9cm}}{\makecell[{{p{9cm}}}]{What has a bed but doesn't sleep and a mouth but never eats?}} \\\cmidrule{1-6}
Candidate & \makecell[c]{ocean \XSolidBrush} &\makecell[c]{pond} &\makecell[c]{river \Checkmark} &\makecell[c]{equator} &\makecell[c]{lake} \\
Prediction probability &\makecell[c]{0.9125} &\makecell[c]{0.0586} &\makecell[c]{0.0167} &\makecell[c]{0.0064} &\makecell[c]{0.0058} \\\midrule
Riddle &\multicolumn{5}{>{\RaggedRight}p{9cm}}{What disappears the moment you say its name?} \\\cmidrule{1-6}
Candidate &\makecell[c]{whisper \XSolidBrush}  &\makecell[c]{darkness}  &\makecell[c]{noise} &\makecell[c]{silence \Checkmark} &\makecell[c]{annihilation} \\
Prediction probability &\makecell[c]{0.4283} &\makecell[c]{0.2516} &\makecell[c]{0.1583} &\makecell[c]{0.1328} &\makecell[c]{0.0291} \\\midrule
Riddle &\multicolumn{5}{>{\RaggedRight}p{9cm}}{\makecell[{{p{9cm}}}]{薄薄一张口，能啃硬骨头。吃肉不喝汤，吃瓜不嚼豆。\\It has a thin mouth and can gnaw bones. It eats meat but doesn't drink soup, eats melons but never munch beans.}} \\\cmidrule{1-6}
Candidate &\makecell[c]{砧板\ \XSolidBrush\\ cutting board} &\makecell[c]{菜刀\ \Checkmark\\ kitchen knife} &\makecell[c]{锅铲\\turner} &\makecell[c]{木棍\\ wooden stick} &\makecell[c]{斧头 \\ axe} \\
Prediction probability &\makecell[c]{0.3568} &\makecell[c]{0.2489} &\makecell[c]{0.1856} &\makecell[c]{0.1622} &\makecell[c]{0.0465}\\\midrule
Riddle &\multicolumn{5}{>{\RaggedRight}p{9cm}}{\makecell[{{p{9cm}}}]{云长想起玄德来\\Yunchang recalled Xuande.}} \\\cmidrule{1-6}
Candidate &\makecell[c]{和蔼可亲\ \XSolidBrush\\ be affable} &\makecell[c]{关怀备至\ \Checkmark\\ care for sb.} &\makecell[c]{以礼相待\\be polite} &\makecell[c]{推心置腹\\ confide in sb.} &\makecell[c]{体贴入微 \\ be considerate} \\
Prediction probability &\makecell[c]{0.2747} &\makecell[c]{0.2210} &\makecell[c]{0.2028} &\makecell[c]{0.1613} &\makecell[c]{0.1402}\\
\bottomrule
\end{tabular}
\caption{Error cases during test by ALBERT-XXL on BiRdQA-en (the upper two) and ERNIE on BiRdQA-zh (the lower two). The cross indicates the candidate chosen by a model, while the checkmark indicates the correct answer. Candidates of each riddle are sorted by their prediction probabilities.}\label{tab:error}
\end{table*}

\section{Comparison with RiddleSense}\label{appendix:comp}
Table \ref{tab:comp-stat} compares some of the key statistics between BiRdQA-en and RiddleSense. BiRdQA-en contains more examples, longer and more diverse questions than RiddleSense, while RiddleSense is more varied in terms of the answers and candidates. 
\begin{table}[H]\centering
\scriptsize
\begin{tabular}{l|cc}\toprule
Measurement &BiRdQA-en &RiddleSense \\\cmidrule{1-3}
\# Total examples &6614 &5715 \\
\# Training examples &4093 &3510 \\
\# Validation examples &1061 &1021 \\
\# Test examples &1460 &1184 \\\cmidrule{1-3}
\# Avg question tokens &29.74 &24.04 \\
Long questions ($>$ 20 tokens) &55.00\% &47.30\% \\
\# Distinct question tokens &15043 &7110 \\
\# Distinct answers &3057 &3622 \\
\# Distinct candidate tokens &6593 &9912 \\
\bottomrule
\end{tabular}
\caption{Comparison of key statistics between BiRdQA-en and RiddleSense.}\label{tab:comp-stat}
\end{table}

Table \ref{tab:comp-result} describes experimental results on BiRdQA-en and RiddleSense. Our dataset has comparable difficulty to RiddleSense. This is non-trivial since we generate distractors \textit{automatically} while the distractors in RiddleSense are collected by \textit{human}. 

\begin{table*}
\centering
\scriptsize
\begin{tabular}{l|cc|cc}\toprule
\multirow{2}{*}{Model} &\multicolumn{2}{c|}{BiRdQA-en} &\multicolumn{2}{c}{RiddleSense} \\\cmidrule{2-5}
&Dev &Test &Dev  &Test \\\cmidrule{1-5}
BERT-Base &48.82 / 0.6834 &41.92 / 0.6404 &54.16 &42.43 \\
BERT-Large &46.65 / 0.6735 &44.25 / 0.6519 &55.24 &45.09 \\
RoBERTa-Large &47.31 / 0.6749 &45.21 / 0.6653 &60.72 &52.58 \\
ALBERT-XXL &63.52 / 0.7856 &58.70 / 0.7590 &\textbf{66.99} &\textbf{60.65} \\
UnifiedQA (T5-Large) &\textbf{67.20} &\textbf{62.60} &56.21 &56.40 \\
\bottomrule
\end{tabular}
\caption{Comparison of results (accuracy/MRR) between BiRdQA-en and RiddleSense. We show the reported dev/test accuracies on RiddleSense \cite{lin-etal-2021-riddlesense} for references.
}\label{tab:comp-result}
\end{table*}

\end{appendices}
\end{CJK*}
\end{document}